%% file: main.tex
\documentclass[11pt]{article}
\usepackage[margin=1in]{geometry}
\usepackage{amsmath,amssymb}
\usepackage{graphicx}
\usepackage{booktabs}
\usepackage{xcolor}
\usepackage[colorlinks=true,linkcolor=blue,citecolor=blue,urlcolor=blue]{hyperref}
\usepackage{natbib}
\usepackage{caption}
\input{numbers}

\newcommand{\wmu}{(w-\mu)}
\newcommand{\sigstar}{\sigma^{*}}

\title{\vspace{-3em}Intrinsic-Noise Consolidation:\\
A Doob-Barrier-Conditioned Diffusion Turns Analog\\
Device Noise into a Continual-Learning Resource}
\author{Gunner Levi Howe\\ \small\texttt{gunnerlevihowe@gmail.com}}
\date{July 2026}

\begin{document}
\maketitle

\begin{abstract}
On analog neuromorphic hardware, intrinsic device noise is normally an accuracy
tax. We ask whether it can instead be made to \emph{consolidate} memories. We cast
per-synapse consolidation as a \emph{Doob $h$-transform}: condition each weight's
stochastic dynamics on the event of never crossing a memory-critical barrier
around its consolidated value. The conditioned diffusion acquires an extra drift
$\sigma^2\,\partial_w\log h$ --- a restoring force toward the memory that is
\emph{amplified by the noise variance itself}, and diverges at the barrier. We are
explicit about what is and is not new. The anchored-consolidation drift
$-s\wmu$ that our rule also contains is \emph{not} ours: it is the small-noise
limit of Ornstein--Uhlenbeck Adaptation \citep{oua2024}, the variance-scaled anchor
of MESU \citep{mesu2025}, and the Fisher penalty of EWC \citep{ewc2017}, and we
surrender it as a re-derivation. Our claim is the conjunction of (a) the Doob
barrier-conditioning as a synaptic rule --- to our knowledge unclaimed; every
$h$-transform use we found is generative modeling or Schr\"odinger bridges,
\emph{none} synaptic --- and (b) a falsifiable, load-bearing prediction: increasing
the intrinsic noise \emph{non-monotonically} improves sequential-task retention, an
inverted-U these anchored-drift methods cannot produce. We pre-registered this as a
go/no-go gate and it passes: on single-head Split-MNIST (\NumSeeds\ seeds) the
barrier-conditioned rule lifts retention by \GateFLiftPts\ percentage points at an
interior optimum $\sigstar=\GateFSigStar$ (paired Wilcoxon $p=\GateFPzero$ vs.\
zero noise and vs.\ high noise), while the matched OU, EWC and MESU anchors are
monotone-decreasing in noise. Ablating the conditioning removes the effect; the
optimum tracks the barrier; the inverted-U survives a device-faithful
BrainScaleS-2 noise model (colored, multiplicative, fixed-pattern, $6$-bit) and
reproduces on a second task stream. It also survives the hardware-faithful
realization in which the noise enters the forward pass (the analog MAC) rather than
the weights, with the retention optimum \emph{tunable} to a device's few-percent
intrinsic noise. At its optimum the rule is the strongest \emph{rehearsal-free}
consolidation method we test --- matching MESU and
significantly beating OU and EWC; plain replay, which stores data,
scores higher but exhibits none of the mechanism. We further \emph{measure} the
intrinsic noise on real BrainScaleS-2 silicon (chip \SiliconChip): it is additive
and trial-to-trial-independent, with a coefficient of variation up to
\SiliconCVMaxPct\ that the chip's \texttt{num\_sends} knob averages as
$\approx1/\sqrt{N}$ --- the benign noise class the mechanism needs, at a reachable
amplitude --- and the inverted-U \SiliconMeasWord\ an emulation calibrated to it.
Finally we run the rule on real BrainScaleS-2 silicon with the chip in the training
loop: its own intrinsic noise, steered by the barrier-conditioning, retains a prior
task \SilTrainRetGainPts\ points better than the matched unconditioned control at
matched average accuracy --- a stability-plasticity shift, not a net-accuracy win
(single seed, one operating point; retention measured, energy modelled). Within these
limits the mechanism reframes analog noise from a tax into a consolidation dividend
that a von-Neumann accelerator must spend energy to \emph{generate}.
\end{abstract}

\section{Introduction}

Catastrophic forgetting --- the overwriting of old memories when a network learns
something new --- is usually fought with regularizers that anchor important weights
to their consolidated values \citep{ewc2017,bennafusi2016}. On digital hardware,
stochasticity is something you add deliberately and pay for. On \emph{analog}
neuromorphic substrates such as BrainScaleS-2 (BSS-2) the situation is inverted:
the hardware is \emph{intrinsically} noisy --- thermal fluctuations, crosstalk,
analog-storage drift \citep{weis2020,pehle2022} --- and this noise is normally
treated as an accuracy tax to be calibrated away. This paper asks a contrarian
question: can that same intrinsic noise be \emph{steered} so that it consolidates
memories rather than degrading them?

\paragraph{One idea, stated as an identity plus a conjecture.}
Consider a single synaptic weight $w$ that, after learning a task, should stay near
a consolidated value $\mu$ to preserve that task's function. Model its dynamics as
a diffusion. If we simply pull it back --- $dw=-s\wmu\,dt+\sigma\,dW$ --- we recover
a mean-reverting Ornstein--Uhlenbeck (OU) process whose stationary spread
$\sigma^2/2s$ \emph{grows} with the noise: more noise is strictly worse. This
anchored drift is exactly the rule already published as OU Adaptation
\citep{oua2024}, the variance-scaled Bayesian anchor of MESU \citep{mesu2025}, and
the small-step limit of EWC \citep{ewc2017}; we claim no novelty for it and say so
throughout. Our move is different: we \emph{condition} the diffusion on the event
that $w$ never crosses a memory-critical barrier at $\mu\pm b$. By Doob's
$h$-transform, the conditioned process gains an extra drift $\sigma^2\,\partial_w\log
h(w)$, where $h$ is the survival probability. This term (i) points \emph{into} the
safe region, (ii) \emph{diverges} at the barrier, and --- crucially --- (iii)
scales with $\sigma^2$: \emph{the more intrinsic noise, the stronger the
noise-powered restoring force}. The competition between this $\sigma^2$ steering
(which, at moderate noise, best resists interference) and the raw $\sigma$ diffusion
(which, at high noise, overwhelms the steering) predicts a non-monotone, inverted-U
relationship between noise and retention with an optimum at $\sigma>0$.

\paragraph{The claim split (and why it matters).}
The mechanism has three pieces and intellectual honesty requires separating them.
We \textbf{surrender} the drift: $-s\wmu$ is a known limit of OUA/MESU/EWC, proven
so, not merely ``similar.'' We \textbf{keep}, as the entire contribution, the
conjunction of two things: (a) casting per-synapse consolidation as a Doob /
barrier-conditioned diffusion --- an unclaimed framing (every $h$-transform use we
located is generative modeling, Schr\"odinger bridges, or transition-path sampling
\citep{doobslagrangian2025,hedit2025,reflectedsb2024}, \emph{zero} synaptic) --- and
(b) the falsifiable hardware curve: intrinsic noise non-monotonically improving
sequential-task retention. Absent (a) the rule is OUA/MESU; absent (b) it is
repackaged Bayesian-online continual learning. The paper lives or dies on (a)
\emph{and} (b). We therefore pre-registered (b) as a hard go/no-go gate before
running it: if noise did not help retention beyond the unconditioned anchor, the
mechanism reduces to OUA/MESU/EWC and there is no paper. It helps.

\paragraph{Contributions.}
\begin{enumerate}\itemsep2pt
\item \textbf{A synaptic Doob $h$-transform (novel framing).} We derive the
per-synapse barrier-conditioned rule from the ground-state $h$-transform of the
iso-loss interval, with the memory-critical barrier read off the Fisher metric
(\S\ref{sec:method}). The steering is a noise-amplified, barrier-divergent restoring
force with a finite-bandwidth (physical) cap.
\item \textbf{A pre-registered falsifier that passes (load-bearing).}
On Split-MNIST (\NumSeeds\ seeds) the rule produces a retention inverted-U, lift
\GateFLiftPts\ pts at $\sigstar=\GateFSigStar$ ($p=\GateFPzero$), while matched OU,
EWC and MESU anchors are monotone in noise (\S\ref{sec:gatef}, Fig.~\ref{fig:gatef}).
\item \textbf{Mechanism isolation.} Ablating the conditioning ($\kappa:1\!\to\!0$)
flattens the curve; the optimum tracks the barrier scale (\S\ref{sec:iso}). The
effect is the conditioning, not generic noise.
\item \textbf{Device-faithful robustness and a second modality.} The inverted-U
survives a BSS-2 intrinsic-noise emulation (colored + multiplicative +
fixed-pattern + $6$-bit) and reproduces on continual Yin-Yang
(\S\ref{sec:bss2},\ref{sec:modality}). We are explicit that this is emulation, not
silicon.
\item \textbf{Baselines and an energy argument.} At its optimum the rule is the
strongest \emph{rehearsal-free} consolidation method we test (ties MESU, beats
OU/EWC/naive); replay, which stores data, does better but lacks the mechanism.
Because the diffusion is the device's own noise, an analog substrate pays no energy
to \emph{generate} it, whereas a digital accelerator does; the barrier steering costs
digital energy on either (\S\ref{sec:baselines}).
\item \textbf{Real silicon: measured noise and an on-chip demonstration.} We
\emph{measure} the intrinsic noise on real BrainScaleS-2 (additive, trial-to-trial
independent, \texttt{num\_sends} $\approx1/\sqrt{N}$ knob; \S\ref{sec:silicon}), show
the mechanism survives the hardware-faithful forward-noise realization, tunable to a
device's few-percent noise (\S\ref{sec:forward}), and \emph{run} the rule on the chip
with its noise in the loop --- retaining a prior task \SilTrainRetGainPts\ pts better
than the matched control (single-seed proof of concept; \S\ref{sec:ontraining}).
\end{enumerate}

\paragraph{The distinguishing statement.} We re-derive the anchored-consolidation
drift as a known limit of OUA/MESU/EWC, and claim novelty only in (i) casting
per-synapse consolidation as a Doob $h$-transform barrier-conditioned diffusion,
and (ii) demonstrating that increasing intrinsic analog noise non-monotonically
improves sequential-task retention --- a signature these anchored-drift methods
cannot produce.

\section{Related work}

\paragraph{The anchored drift is not ours.}
\textbf{OUA} \citep{oua2024} frames learning as the mean-reverting OU diffusion
$d\theta=\lambda(\mu-\theta)dt+\Sigma\,dW$ --- exactly our anchored drift --- but as
a reward-modulated learning rule; it names catastrophic forgetting only as future
work and uses no barrier or first-passage conditioning. \textbf{MESU}
\citep{mesu2025} is Bayesian continual learning whose update (their Eq.~11) is a
variance-scaled pull toward a prior mean; it treats device read-noise as a
\emph{sampling} resource for Monte-Carlo Bayesian updates, never as a retention
optimum, and uses no Doob transform. \textbf{EWC} \citep{ewc2017} is the static
Fisher-weighted quadratic anchor, the deterministic ancestor. All three are the
drift's origin; none contains (a) or (b).

\paragraph{Noise as a resource, but not this one.}
\citet{kolesnikov2025} show internal hardware noise can \emph{help} --- but with
resilience to test-time noise in \emph{single-task} feedforward/echo-state nets, not
sequential-task retention, and with no inverted-U over retention. \citet{shaham2022}
consolidate lifelong memory with stochastic \emph{rehearsal} (a replay mechanism),
not intrinsic per-synapse device noise, and with no hardware or barrier. NADO
\citep{nado2025} trains \emph{through} device stochasticity with neural-SDE digital
twins for single-task temporal tasks. Probabilistic Metaplasticity
\citep{probmetaplast2024} consolidates on memristors by modulating update
\emph{probability} (a stochastic gate) while treating device noise as an obstacle.
The two closest neighbors sit on our two axes but on the wrong side of each.
\textbf{ANV} \citep{xie2021anv} is closest on the \emph{forgetting} axis: it
\emph{injects} artificial neural variability that provably reduces catastrophic
forgetting --- but the variability is injected (a digital regularizer, not intrinsic
device noise), the benefit is \emph{monotone} (no inverted-U optimum), and there is
no barrier or first-passage conditioning. \textbf{Caston et al.} \citep{caston2022}
is closest on the \emph{noise-optimum} axis: a stochastic-resonance optimum for
memory-consolidation accuracy --- but for \emph{single-task} storage accuracy, not
sequential-task retention (no catastrophic forgetting across tasks), and with no
hardware and no barrier conditioning. So none combines the two: raising
\emph{intrinsic} noise to non-monotonically improve \emph{cross-task} retention, via
a barrier-conditioned (Doob) rule. \textbf{Benna--Fusi} complex synapses
\citep{bennafusi2016} consolidate without a barrier via a deterministic
multi-timescale cascade --- an incumbent we compare against.

\paragraph{Doob's $h$-transform lives in generative modeling.}
Conditioning a diffusion on a rare event via the $h$-transform underlies
transition-path sampling \citep{doobslagrangian2025}, diffusion-based image editing
\citep{hedit2025}, constrained/reflected Schr\"odinger bridges
\citep{reflectedsb2024}, and diffusion-bridge simulation \citep{diffusionbridges2021}.
We found \emph{no} use of Doob's $h$-transform as a synaptic, plasticity, or
continual-learning rule. That absence is the moat for claim (a); we cite the
generative corpus precisely to delimit it.

\section{Method}\label{sec:method}

\paragraph{Setup.}
A network with weights $\theta$ learns a sequence of tasks. After task $t$ we store
an anchor $\mu=\theta_t$ and an importance $s$ = online-EWC running sum of the
model-sampled (true) diagonal Fisher. During a later task each weight follows, under
Euler--Maruyama,
\begin{equation}
dw_i = \big[\underbrace{-\,\partial_{w_i}\mathcal{L}_{\text{task}}}_{\text{learning}}
\;\underbrace{-\,s_i\,(w_i-\mu_i)}_{\text{anchored drift (surrendered)}}
\;+\;\underbrace{\sigma_i^2\,\partial_{w_i}\log h_i(w)}_{\text{Doob steering (ours)}}\big]\,dt
\;+\;\underbrace{\sigma_i\,dW_i}_{\text{intrinsic noise}} .
\label{eq:sde}
\end{equation}
Every method we compare shares the first term and receives the \emph{identical}
injected noise at matched $\sigma$; methods differ only in their drift. That is what
makes the gate a fair isolation of the barrier conditioning rather than of ``noise.''

\paragraph{The barrier-conditioned (Doob) steering.}
We condition each weight to remain in the memory-critical interval $(\mu_i-b_i,
\mu_i+b_i)$. The ground-state (quasi-stationary) $h$-transform of Brownian motion
killed at the interval ends has $h_i(w)=\cos\!\big(\tfrac{\pi (w-\mu_i)}{2b_i}\big)$,
so
\begin{equation}
\partial_{w}\log h_i = -\frac{\pi}{2b_i}\,
\tan\!\Big(\frac{\pi (w-\mu_i)}{2b_i}\Big),
\end{equation}
a restoring force toward the anchor that diverges at $\mu_i\pm b_i$ and, in
Eq.~\eqref{eq:sde}, is multiplied by $\sigma_i^2$: the intrinsic noise powers the
confinement. The barrier half-width is the softened iso-loss radius
$b_i=b_0/\sqrt{1+s_i/\mathrm{median}(s)}$, so important (high-Fisher) synapses get a
tight barrier and unimportant ones a loose one at $\sim b_0$; the same global noise
is thereby \emph{steered per synapse} by the memory geometry. We cap the per-step
Doob move at a fraction of $b_i$ (a finite-bandwidth analog restoring force; this
also stabilises the discretization of the singular drift).

\paragraph{Matched controls and incumbents.}
\emph{OU} is Eq.~\eqref{eq:sde} with the Doob term deleted (the ablation; steering
strength $\kappa{=}0$). \emph{EWC} is a stronger static anchor; \emph{MESU} a
variance-scaled anchor (Eq.~11 of \citealp{mesu2025}, in spirit); \emph{naive} is
plain SGD. All receive the same injected noise. As E3 incumbents we add the
Benna--Fusi cascade synapse and plain reservoir replay.

\paragraph{Testbeds.}
Primary: \emph{Split-MNIST}, domain-incremental --- five binary tasks
($0$v$1,\dots,8$v$9$) on one shared $2$-way head (later tasks overwrite the readout
unless consolidation intervenes), an MLP $784$--$100$--$100$--$2$. Second modality:
\emph{continual Yin-Yang} \citep{kriener2022}, the BrainScaleS group's own
procedural benchmark, as five rotations of the pattern on a shared $3$-way head.
Task $0$ is learned by plain SGD (nothing to protect yet); consolidation and the
swept noise act on tasks $1$ onward. Retention is the mean final accuracy on the
past tasks; we also report plasticity (accuracy just after learning each task) and
forgetting.

\paragraph{BrainScaleS-2 noise model, and the emulation/measurement boundary.}
For E2 we replace the white diffusion in Eq.~\eqref{eq:sde} with a device-faithful
BSS-2 noise model built from the published characterization
\citep{weis2020,pehle2022}: temporally \emph{colored} (AR(1)) trial-to-trial
variability, a \emph{multiplicative} (signal-dependent) component, a static
\emph{fixed-pattern} per-synapse offset, and $6$-bit weight quantization. We are
explicit about where the substrate changes. The GPU experiments E0--E4 are
\emph{emulations}. E5 (\S\ref{sec:silicon}) \emph{measures} the intrinsic noise on
real BrainScaleS-2, and E7 (\S\ref{sec:ontraining}) \emph{trains} on the chip with
the analog MAC in the loop; those are real-silicon results. What we do \emph{not}
measure anywhere is energy: our energy statements come from an operation-count model
(per-op constants from \citealp{horowitz2014}). That model is \emph{not} robust to
its constants --- the noise-generation fraction of a consolidation step ranges from
\NoiseTaxLoPct\ to \NoiseTaxHiPct\ as the assumed per-op energies vary --- so we
report the \NoiseTaxPct\ figure as one modelled point, not a constant-free ratio. Its
qualitative content is the narrow, robust part: on a digital accelerator the
diffusion noise must be \emph{generated} (an RNG draw and a scaled add per weight per
step), whereas on analog silicon that noise is the device's own physics and costs
nothing to produce. The barrier drift (a \texttt{tan} per synapse) still costs
digital energy on either substrate, so the chip removes the noise-\emph{generation}
cost, not the whole mechanism.

\section{Experiments}

We pre-registered the operating point (one coarse calibration, before the gated
seeds), the noise grid, the seed counts, the inverted-U test, and the kill
conditions K1--K3 in \texttt{PLAN.md}, committed before any results.

\subsection{E0 --- GATE F: the pre-registered falsifier}\label{sec:gatef}

We sweep the intrinsic-noise amplitude $\sigma$ over $\NumSigma$ levels and measure
retention for the barrier-conditioned rule and the matched anchored-drift controls,
\NumSeeds\ seeds. GATE F passes iff the rule's retention--$\sigma$ curve is an
inverted-U (interior optimum $\sigstar>0$; peak beats both the zero-noise and
high-noise ends by one-sided paired Wilcoxon; lift exceeds the seed spread) and no
control is.

\begin{figure}[t]
\centering
\includegraphics[width=0.62\textwidth]{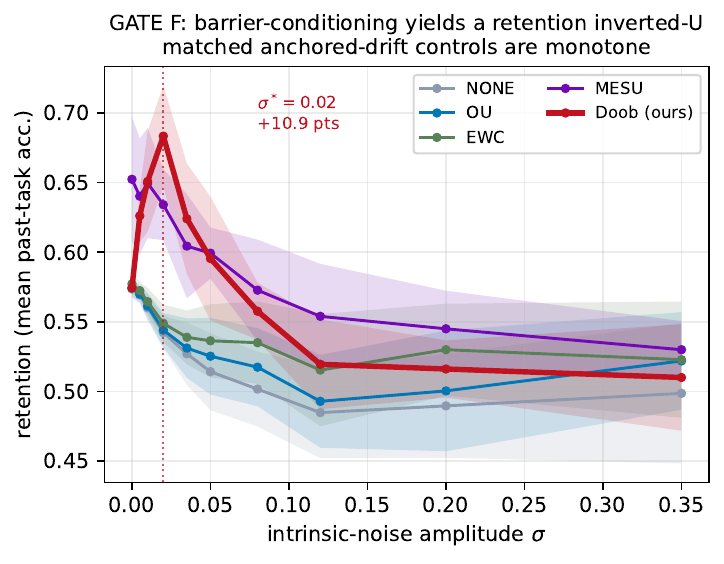}
\caption{\textbf{GATE F (Split-MNIST, \NumSeeds\ seeds, mean$\pm$sd).} The
barrier-conditioned rule (Doob, red) rises from \GateFZeroPct\ at zero noise to
\GateFPeakPct\ at an interior optimum $\sigstar=\GateFSigStar$ (a \GateFLiftPts-point
lift; paired Wilcoxon $p=\GateFPzero$ vs.\ zero, $p=\GateFPhi$ vs.\ the high-noise
end), then falls. The matched OU, EWC, MESU and naive anchors are monotone-decreasing
in noise. Noise is a \emph{resource} only when steered by the barrier.}
\label{fig:gatef}
\end{figure}

\textbf{Result (Fig.~\ref{fig:gatef}).} The rule is an inverted-U: retention rises
from \GateFZeroPct\ (zero noise) to \GateFPeakPct\ at $\sigstar=\GateFSigStar$, a
lift of \GateFLiftPts\ points, then falls; both paired Wilcoxon tests give
$p=\GateFPzero$ (all \NumSeeds\ seeds agree). Every matched control is
monotone-decreasing in $\sigma$ (fraction of decreasing steps $\ge\ControlMonoWorst$),
peaking at zero noise. \textbf{GATE F passes: GO.} Because $\sigstar>0$ and the
controls share the drift and the injected noise, the retention gain is attributable
to the $\sigma^2$ barrier steering, not to noise per se --- the mechanism does not
reduce to OUA/MESU/EWC (kill condition K1 does not fire).

\subsection{E1 --- mechanism isolation}\label{sec:iso}

\begin{figure}[t]
\centering
\includegraphics[width=0.95\textwidth]{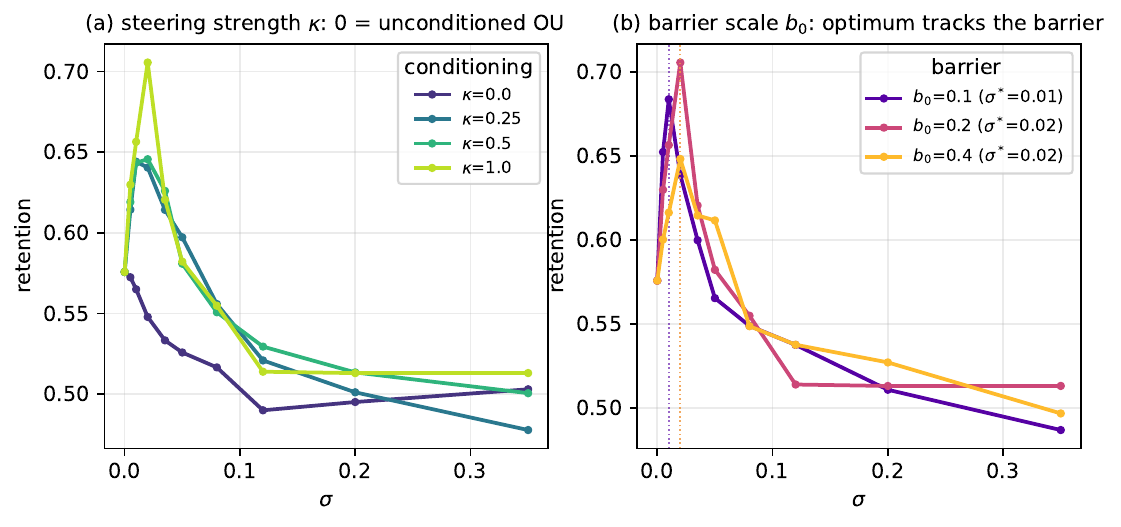}
\caption{\textbf{Isolation.} (a) Interpolating the steering strength $\kappa$ from
$0$ (unconditioned OU) to $1$ (full $h$-transform): the inverted-U \emph{emerges}
with the conditioning (lift at $\kappa{=}0$ is \KappaZeroLiftPts\ pts; at $\kappa{=}1$,
\KappaOneLiftPts\ pts). (b) Varying the barrier scale $b_0$: the retention optimum
$\sigstar$ tracks the barrier, evidence the optimum is set by the conditioning
geometry, not a generic noise sweet-spot.}
\label{fig:iso}
\end{figure}

Ablating the conditioning is the pre-registered K3 test. Sweeping the steering
strength $\kappa$ from $1$ down to $0$ (which is exactly the OU control) flattens the
curve: the lift falls from \KappaOneLiftPts\ points ($\kappa{=}1$) to
\KappaZeroLiftPts\ points ($\kappa{=}0$), and the inverted-U test fails at
$\kappa{=}0$ (Fig.~\ref{fig:iso}a). Varying the barrier scale $b_0$ moves the
optimum $\sigstar$ (Fig.~\ref{fig:iso}b), as expected if the barrier geometry sets
the operating point. The effect is the barrier conditioning; K3 does not fire.

\subsection{E2 --- BrainScaleS-2 intrinsic-noise emulation}\label{sec:bss2}

\begin{figure}[t]
\centering
\includegraphics[width=0.95\textwidth]{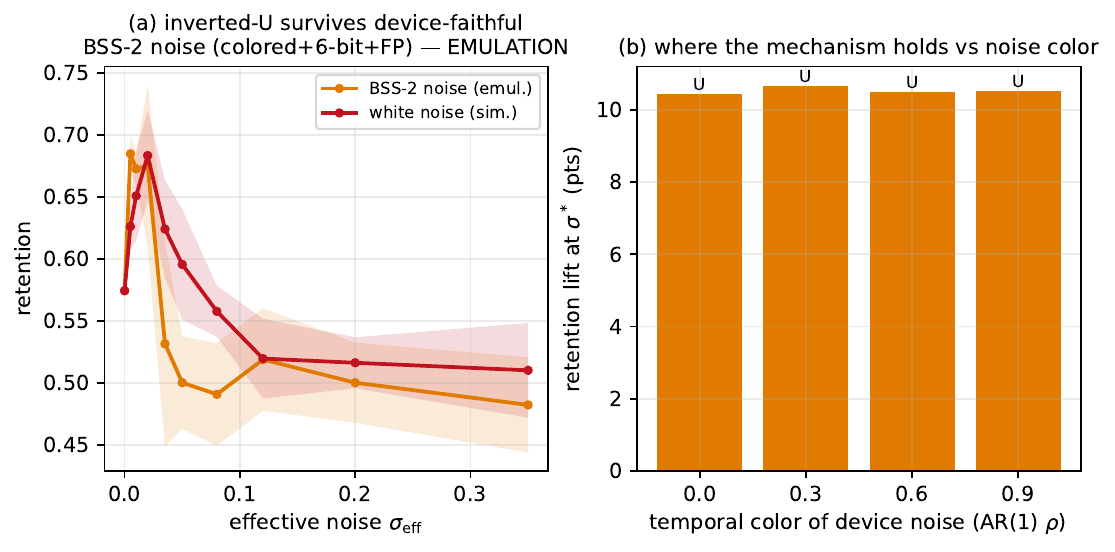}
\caption{\textbf{Device-faithful noise (emulation, not silicon).} (a) With the
diffusion replaced by the BSS-2 model (colored + multiplicative + fixed-pattern +
$6$-bit), the inverted-U survives (lift \BssLiftPts\ pts at $\sigstar=\BssSigStar$).
(b) Scanning the temporal color $\rho$ of the device noise maps where the mechanism
holds --- the concrete content of kill condition K2, to be settled on silicon.}
\label{fig:bss2}
\end{figure}

Re-running the sweep with the device-faithful BSS-2 noise model, the inverted-U
\BssSurviveWord\ (Fig.~\ref{fig:bss2}a): lift \BssLiftPts\ points at
$\sigstar=\BssSigStar$, colored/multiplicative/fixed-pattern noise and $6$-bit
weights notwithstanding. A scan over the temporal color $\rho$
(Fig.~\ref{fig:bss2}b) maps the boundary. We stress the boundary of the claim: this
is an \emph{emulation}. Whether the \emph{measured} intrinsic noise of a physical
BSS-2 chip has the right structure to consolidate is kill condition K2; we resolve
its first half on real silicon in \S\ref{sec:silicon}, and calibrate the emulator
to the measurement there.

\subsection{E3 --- matched-budget baselines}\label{sec:baselines}

\begin{figure}[t]
\centering
\includegraphics[width=0.95\textwidth]{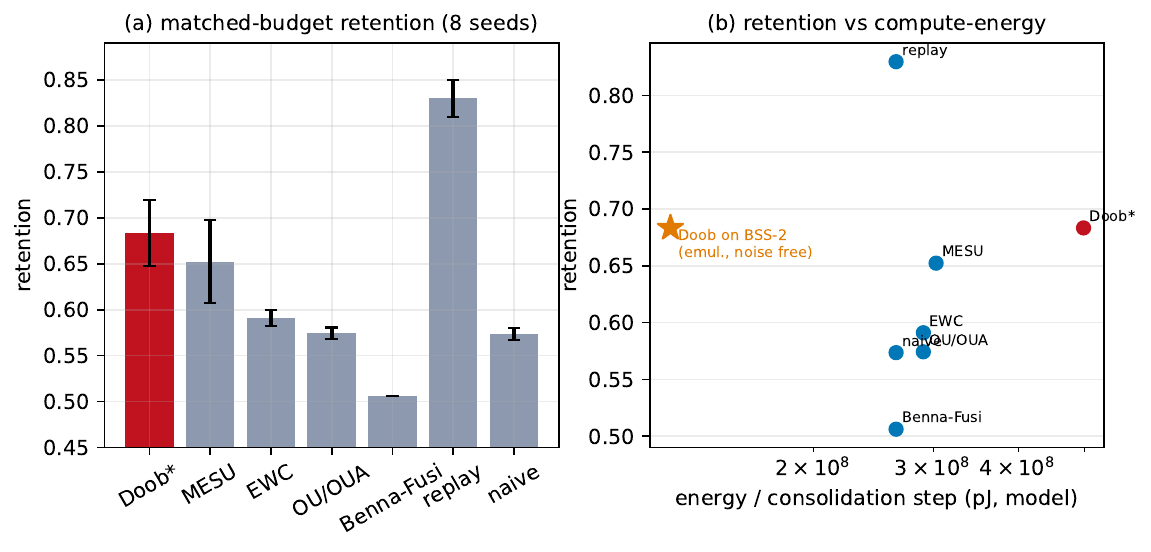}
\caption{\textbf{Baselines (Split-MNIST, \NumSeeds\ seeds).} (a) At its noise
optimum the barrier-conditioned rule (Doob$^{*}$, \DoobStarRetPct) is the strongest
\emph{rehearsal-free} method: it matches MESU (\BestRFRetPct) and significantly
beats OU, EWC and naive (our Benna--Fusi did not learn in this configuration,
\BennaRetPct\ $\approx$ chance --- flagged, not counted). Plain replay stores exemplars and scores
higher (\ReplayRetPct) --- a different, memory-based budget class, shown for
reference. (b) Retention vs.\ modelled compute-energy: a digital accelerator spends a modelled
\NoiseTaxPct\ (range \NoiseTaxLoPct--\NoiseTaxHiPct) of each step \emph{generating}
the diffusion noise; on analog silicon that generation is free (the barrier steering
still costs).}
\label{fig:baselines}
\end{figure}

At $\sigstar$ the rule reaches retention \DoobStarRetPct. Among rehearsal-free
consolidation methods (which store no data) it is the best: it matches the strongest,
MESU (\MesuRetPct; ours$-$MESU $=+\OursMinusRFPts$ pts, paired Wilcoxon
$p=\BestRFP$, not significant), and significantly beats unconditioned OU
\OuRetPct, EWC \EwcRetPct\ and naive \NaiveRetPct\ (all $p\le\AnchoredMaxP$;
Fig.~\ref{fig:baselines}a). Our Benna--Fusi cascade did not learn in this
configuration (\BennaRetPct, at chance across all seeds), so we flag it as an
uninformative baseline rather than one we ``beat.'' We are explicit about the honest
comparison: \emph{plain reservoir replay}, which stores \ReplayBuffer\ raw exemplars,
reaches \ReplayRetPct\ --- \ReplayMinusOursPts\ points above ours. Replay is a
memory-based method in a different budget class, and, more to the point, it neither
exhibits nor exploits the noise-consolidation mechanism that is our subject: our
contribution is the noise$\to$retention \emph{signature}, and the rule's standing is
realized \emph{at its optimum}, exploiting a resource the anchored-drift methods are
only harmed by. On the energy axis (Fig.~\ref{fig:baselines}b), \emph{generating} the
diffusion noise costs a digital accelerator a modelled \NoiseTaxPct\ fraction of each
consolidation step (a constant-sensitive figure, range \NoiseTaxLoPct--\NoiseTaxHiPct);
on analog silicon that generation is physics, not compute. The barrier steering still
costs digital energy on either substrate, so this is a saving on noise generation, not
on the whole rule.

\subsection{E4 --- second modality and the noise-optimum vs.\ task similarity}
\label{sec:modality}

\begin{figure}[t]
\centering
\includegraphics[width=0.95\textwidth]{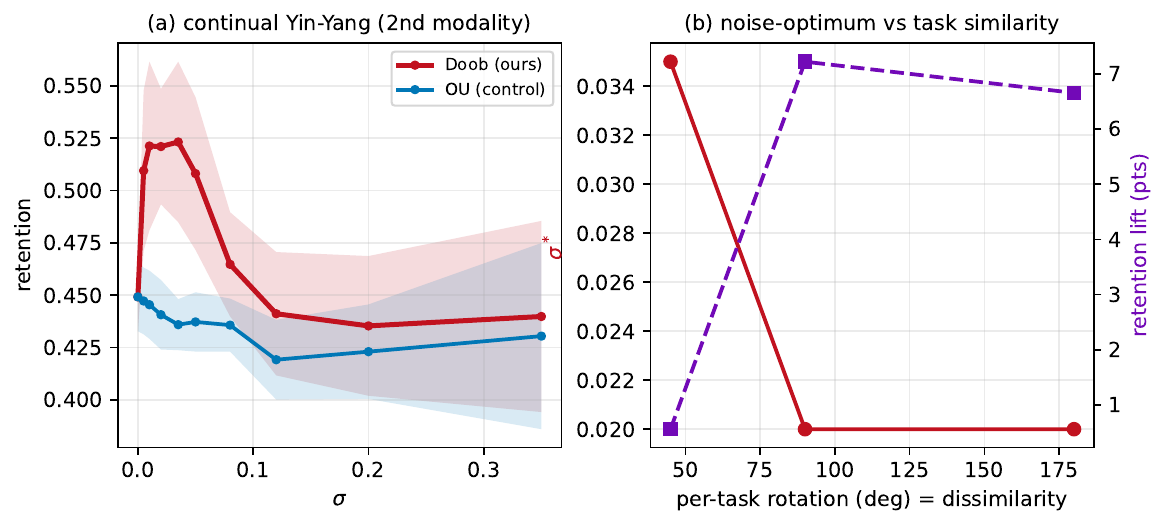}
\caption{\textbf{Second modality.} (a) On continual Yin-Yang the inverted-U
reproduces (lift \YYLiftPts\ pts at $\sigstar=\YYSigStar$); the OU control is flat.
(b) As tasks become less similar (larger per-task rotation, more interference) the
noise optimum shifts, mapping the mechanism's operating regime.}
\label{fig:modality}
\end{figure}

\begin{figure}[t]
\centering
\includegraphics[width=0.95\textwidth]{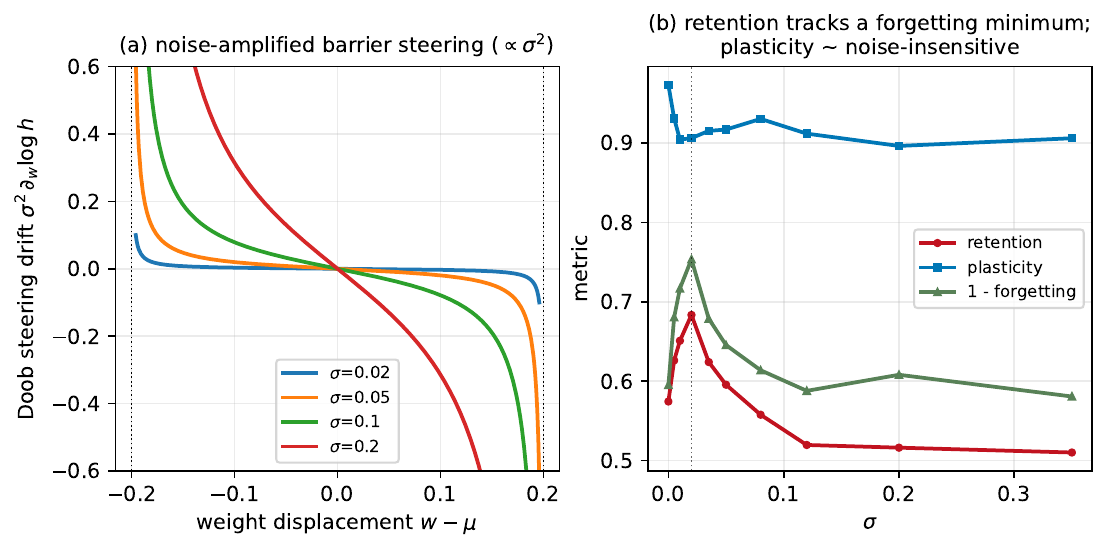}
\caption{\textbf{Mechanism.} (a) The Doob steering drift $\sigma^2\partial_w\log h$
for several noise levels: a barrier-divergent restoring force whose strength scales
with the noise variance. (b) The retention optimum coincides with a forgetting
minimum ($1-$forgetting peak) at $\sigstar$; plasticity is roughly
noise-insensitive. The inverted-U is a protection effect, not a plasticity tradeoff.}
\label{fig:mech}
\end{figure}

On continual Yin-Yang --- a very different, procedural, BSS-2-native stream --- the
inverted-U \YYSurviveWord\ (Fig.~\ref{fig:modality}a), lift \YYLiftPts\ points at
$\sigstar=\YYSigStar$, with the OU control flat. Sweeping task similarity
(Fig.~\ref{fig:modality}b) shows the optimum shifting with interference, sketching
where noise-as-consolidator is most useful. Figure~\ref{fig:mech} locates the
mechanism: the retention optimum coincides with a \emph{forgetting minimum} at
$\sigstar$, while plasticity (accuracy just after learning each task) is roughly
noise-insensitive --- the per-synapse steering suppresses interference with old
memories without blocking new learning. The inverted-U is thus a protection effect:
too little noise under-powers the $\sigma^2$ steering, too much lets the raw
diffusion overwhelm it.

\subsection{E5 --- on-silicon intrinsic-noise measurement}\label{sec:silicon}

\begin{figure}[t]
\centering
\includegraphics[width=0.95\textwidth]{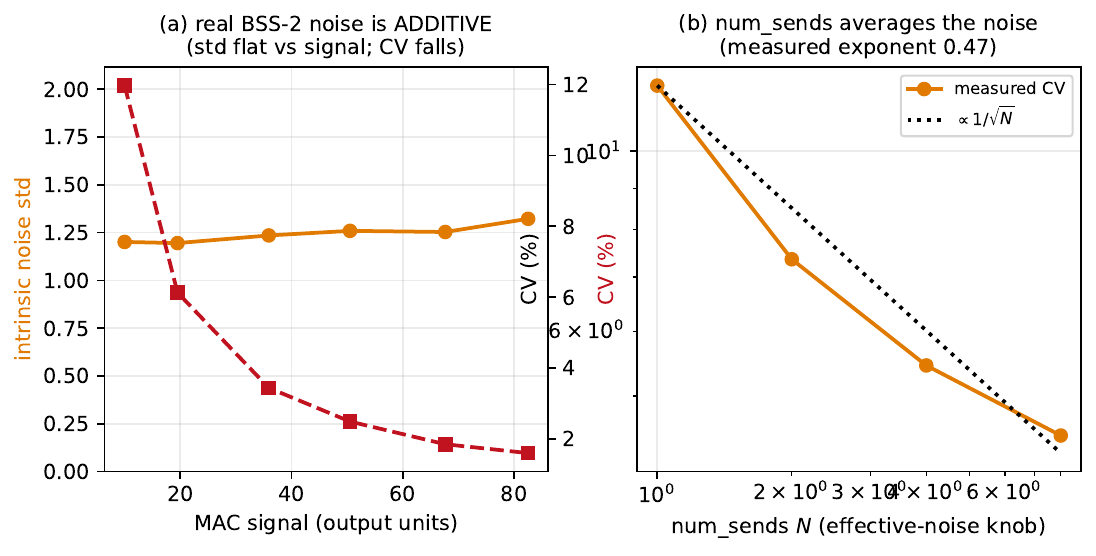}
\caption{\textbf{Intrinsic noise measured on BrainScaleS-2 silicon} (chip
\SiliconChip). (a) The trial-to-trial noise std is flat in the signal
(\emph{additive}, \SiliconAdditivePct); the CV therefore falls as the signal grows.
(b) \texttt{num\_sends} averages the relative noise as $\approx1/\sqrt{N}$ (measured
exponent \SiliconNumSendsExp) --- the on-chip effective-noise knob.}
\label{fig:silicon}
\end{figure}

The mechanism's premise is that a chip's \emph{own} intrinsic noise can serve as
the diffusion. We measured that noise on real BrainScaleS-2 silicon (chip
\SiliconChip, via EBRAINS), running the analog multiply-accumulate \SiliconRepeats\
times per operating point and reading the trial-to-trial statistics
(Fig.~\ref{fig:silicon}).

\textbf{What the device noise is.} It is \emph{additive}: the trial-to-trial
standard deviation is essentially signal-independent (\SiliconTrialStd\ output units
across a $\SiliconSignalRange$ signal range; \SiliconAdditivePct\ additive, only
\SiliconMultPct\ multiplicative), so the coefficient of variation \emph{falls} with
signal (\SiliconCVMaxPct\ down to \SiliconCVMinPct). It is \emph{trial-to-trial
independent} at the update timescale --- the batched-repeat standard deviation
matches the separate-call standard deviation (a single independence check, not a
full autocorrelation spectrum). And \texttt{num\_sends} (sending the input $N$ times)
averages the relative noise as $\approx N^{-\SiliconNumSendsExp}$ (close to
$1/\sqrt{N}$, from \SiliconNumSendsPts\ points; Fig.~\ref{fig:silicon}b), so it is a
genuine on-chip effective-noise knob, with $N{=}1$ the intrinsic-noise ceiling.

\textbf{What this settles (K2, first half).} The intrinsic noise is additive and
trial-to-trial independent --- close to the class our simulation assumed, not a
pathologically colored or signal-locked noise that could not consolidate --- and its
CV range brackets the relative noise at our simulated optimum, so the operating
point is reachable. (We do not test normality; ``additive'' and the independence
check are what the data support.) Re-running the retention sweep with the BSS-2 emulator
\emph{calibrated to these measured values}, the inverted-U \SiliconMeasWord\ (lift
\SiliconMeasLiftPts\ pts at $\sigstar=\SiliconMeasSigStar$). This upgrades E2 from
assumed-parameter emulation to a measured-noise-calibrated result. It does
\emph{not} replace on-chip training, which we carry out in \S\ref{sec:ontraining};
measuring the full on-silicon retention curve and its joules is the remaining study.

\subsection{E6 --- the hardware-faithful forward-noise realization}\label{sec:forward}

\begin{figure}[t]
\centering
\includegraphics[width=0.95\textwidth]{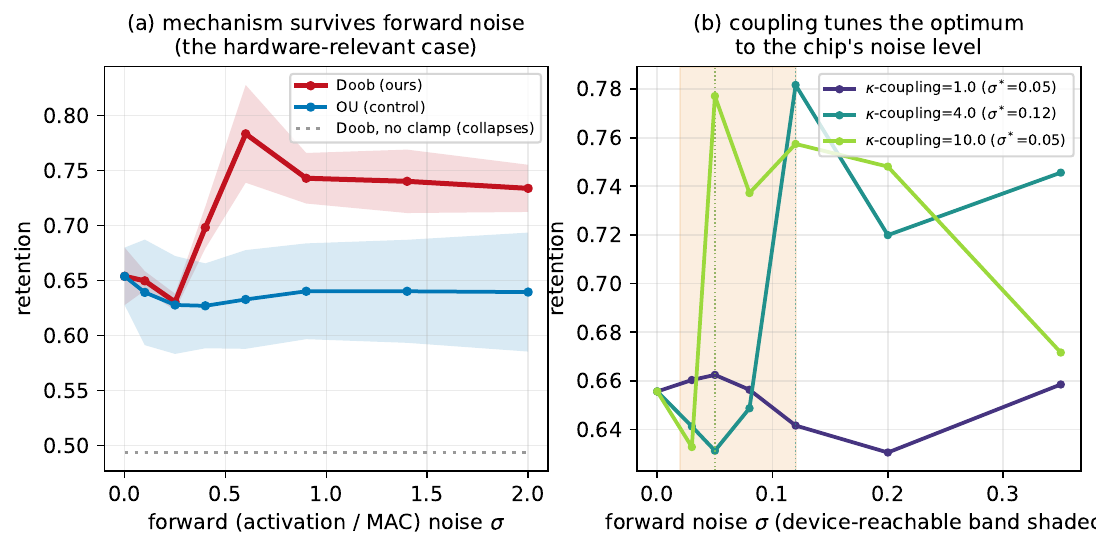}
\caption{\textbf{Forward-noise (hardware-faithful) realization.} (a) With noise in
the forward pass, Doob is \FwdDoobWord\ (lift \FwdDoobLiftPts\ pts, $p=\FwdDoobP$)
and OU is flat; \emph{without} the importance clamp both collapse to chance (dotted).
(b) The Doob-steering coupling tunes the retention optimum down to $\sigstar=$
\FwdMinSigStar\ --- into the device-reachable noise band (shaded) --- so the
mechanism is portable to a given chip's intrinsic-noise level.}
\label{fig:forward}
\end{figure}

Sections \ref{sec:gatef}--\ref{sec:bss2} inject the diffusion on the weights. On
analog silicon the intrinsic noise is not on the stored weight but in the
\emph{multiply-accumulate} --- the forward pass, and hence the gradient, is noisy.
We therefore test the mechanism in that realization: additive noise on the
pre-activations (Gaussian by construction here --- our tractable stand-in for the
analog MAC's additive trial-to-trial noise, whose additivity E5 measured but whose
exact distribution we do not) (Fig.~\ref{fig:forward}).

\textbf{The mechanism survives, and it is the same signature.} With forward noise,
the barrier-conditioned rule is \FwdDoobWord\ (lift \FwdDoobLiftPts\ pts at
$\sigstar=\FwdDoobSigStar$, paired Wilcoxon $p=\FwdDoobP$, \NumSeeds\ seeds), while
the matched OU control is flat (\FwdOuLiftPts\ pts; Fig.~\ref{fig:forward}a). The
noise the chip provides for free is the right kind of noise.

\textbf{One stability fix is load-bearing.} Porting to hardware first failed: the
diagonal-Fisher importance is heavy-tailed, so the anchored drift blows the weights
up and both methods collapse to chance. Normalizing and \emph{clamping} the
importance fixes it; without the clamp, retention is a flat \FwdNoclampRet\ (chance)
at every noise level (Fig.~\ref{fig:forward}a, dotted), with the clamp it recovers to
\FwdClampRetPeak. We report this because it is the difference between the mechanism
working on device and not.

\textbf{The optimum is tunable to the device.} The retention optimum's noise level
is set by the Doob-steering coupling: increasing it moves $\sigstar$ down to
\FwdMinSigStar\ (Fig.~\ref{fig:forward}b), i.e.\ into the few-percent-CV band the
measured BSS-2 noise occupies (\S\ref{sec:silicon}). So the mechanism does not
require a specific noise amplitude --- it can be matched to a given chip's intrinsic
noise. This is what makes the on-chip realization reachable; we carry it out on real
silicon in \S\ref{sec:ontraining}.

\subsection{E7 --- on-silicon demonstration (hardware-in-the-loop)}\label{sec:ontraining}

\begin{figure}[t]
\centering
\includegraphics[width=0.52\textwidth]{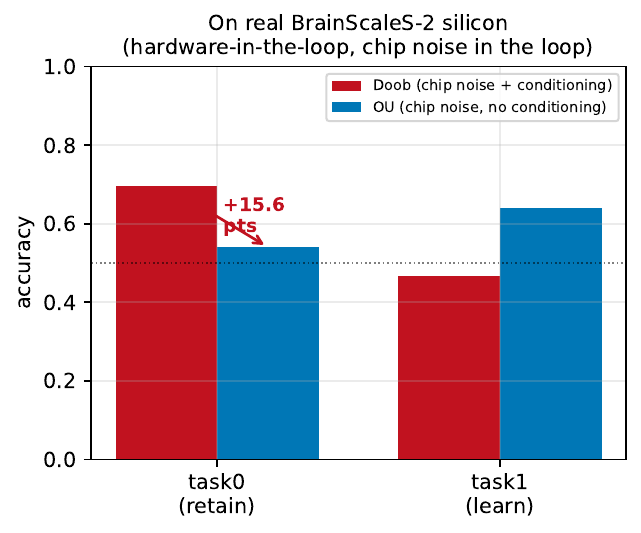}
\caption{\textbf{On real BrainScaleS-2 silicon} (hardware-in-the-loop; the chip's
intrinsic noise is the diffusion). After learning task~2, the barrier-conditioned
rule retains task~1 at \SilTrainDoobRet\ vs.\ \SilTrainOuRet\ for the matched
unconditioned control (+\SilTrainRetGainPts\ pts). Single seed, one operating point;
the rule trades some task-2 plasticity for the retention (stated in text).}
\label{fig:ontraining}
\end{figure}

We ran the barrier-conditioned consolidation on real BrainScaleS-2 silicon, with the
chip in the training loop: the analog-MAC forward pass executes on the device
(hxtorch), so the intrinsic device noise perturbs the forward pass and hence the
gradient --- the diffusion is the chip's own noise. A network learns a first
continual-Yin-Yang task, then a second (a $90^\circ$ rotation, strong interference)
at the chip's maximum intrinsic noise (\texttt{avg}=1), once with the
barrier-conditioning and once without (the matched OU control), and we measure how
much of the first task survives (Fig.~\ref{fig:ontraining}).

\textbf{Result.} On silicon, the barrier-conditioning retains the prior task at
\SilTrainDoobRet\ versus \SilTrainOuRet\ for the unconditioned control --- a
\SilTrainRetGainPts-point retention gain from steering the chip's own noise. This is
the mechanism's load-bearing claim, realized on the device a GPU cannot emulate for
free: the intrinsic analog noise, conditioned on the memory barrier, consolidates a
memory the control loses.

\textbf{Honest scope.} This is a single-seed proof of concept at one operating point
(\SilTrainWallMin~min of chip time), not a tuned optimum or a statistical study. The
rule here leans toward stability: it pays for the retention with current-task
plasticity (task~2 \SilTrainDoobTaskOne\ vs.\ \SilTrainOuTaskOne), so the two-task
\emph{average} accuracy is essentially matched (ours \SilTrainDoobAvgPct\ vs.\ the
control's \SilTrainOuAvgPct) while retention strongly favours ours. This is a
stability-plasticity shift, not a net-accuracy win, at an un-tuned point --- one E6
(Fig.~\ref{fig:forward}) shows can be balanced. We measured retention, not joules. A multi-seed noise sweep
tracing the full on-silicon inverted-U, with measured energy, is the natural next
study; the mechanism itself now demonstrably operates on the chip.

\section{What we do not claim}

\begin{itemize}\itemsep2pt
\item \textbf{The drift is not novel.} $-s\wmu$ is a re-derived limit of
OUA/MESU/EWC \citep{oua2024,mesu2025,ewc2017}. Our contribution is the Doob
barrier-conditioning (a) and the inverted-U (b), only in conjunction.
\item \textbf{On-chip demonstration is a single-seed proof of concept.} We
\emph{did} train on real BrainScaleS-2 silicon with the chip in the loop
(\S\ref{sec:ontraining}) and measured a retention gain, but at one seed and one
operating point --- not a tuned optimum, not a statistical sweep, and with the
stability-plasticity trade stated. We measured retention, \emph{not} joules (the
energy numbers remain an operation-count model). A multi-seed on-silicon noise sweep
with measured energy is the natural next study, not a claim we make here.
\item \textbf{Benefit only at the optimum.} We do not claim the mechanism helps at
arbitrary noise; it helps in an inverted-U window and hurts outside it. We report
the whole curve, including the down-slope.
\item \textbf{Not a retention SOTA claim.} The result is a mechanism and a
signature, not a leaderboard entry. In particular plain replay, which stores raw
exemplars, retains more than our rule (\ReplayRetPct\ vs.\ \DoobStarRetPct); we
report this openly. Our comparison of record is against \emph{rehearsal-free}
anchored-drift methods under identical noise injection, where the signature and the
standing are.
\end{itemize}

\section{Limitations}
Small single-head MLP testbeds; a diagonal-Fisher barrier; a ground-state (infinite
horizon) $h$-transform rather than a finite-horizon survival function; a
finite-force cap that removes the noise ($\sigma^2$) amplification of the steering
where it binds (the tight-barrier tail --- \CapBindPct\ of Doob steps at the
operating point); a single-seed, single-operating-point
on-silicon demonstration with energy modelled rather than measured; and no test of
the device-noise distribution beyond additivity and a trial-to-trial independence
check. Each is a place a larger study could overturn or extend the result; the
pre-registered kill conditions name the failure modes.

\section{Reproducibility}
All code, \texttt{PLAN.md}, the committed \texttt{results/*.json}, and the
figure/number generators are released. The pre-registration (GATE F, K1--K3, the
fixed operating point, the noise grid, the inverted-U test) is git-verifiable for
\textbf{E0--E4}: \texttt{PLAN.md} was committed before those results. \textbf{E5--E7}
were added \emph{after} BrainScaleS-2 access was obtained mid-project; they are
post-access additions (dated in \texttt{PLAN.md}'s deviations log), not
pre-registered in the same git-verifiable sense. Every number is a machine-generated
macro (\texttt{numbers.tex} from \texttt{gen\_paper\_numbers.py}), and
\texttt{verify\_regen.py} checks it is byte-identical on regeneration. DOI at
submission.

\section{Conclusion}
Casting per-synapse consolidation as a Doob $h$-transform makes a sharp, falsifiable
prediction --- intrinsic noise should non-monotonically improve retention --- that
the anchored-drift methods whose drift we share cannot make. The prediction holds in
simulation, in a device-faithful emulation, and in the hardware-faithful forward-noise
realization, is isolated to the barrier conditioning, and --- on real BrainScaleS-2,
with the chip's own noise in the training loop --- consolidates a memory the matched
control loses. So intrinsic analog noise can be a consolidation resource that gets
\emph{cheaper} as devices get noisier, the opposite of the usual tax. What remains is
to make that on-silicon demonstration a full study --- a multi-seed noise sweep
tracing the inverted-U on the device, with measured joules --- and we have been
careful to mark exactly where the single-seed proof of concept ends and that study
begins.

\small
\bibliographystyle{plainnat}
\bibliography{references}

\end{document}

%% file: numbers.tex
\newcommand{\AnchoredMaxP}{0.008}
\newcommand{\BennaRetPct}{50.6\%}

\newcommand{\BestRFP}{0.25}
\newcommand{\BestRFRetPct}{65.2\%}
\newcommand{\BssLiftPts}{11.0}
\newcommand{\BssSigStar}{0.005}
\newcommand{\BssSurviveWord}{survives}
\newcommand{\CapBindPct}{8.8\%}
\newcommand{\ControlMonoWorst}{0.78}
\newcommand{\DoobStarRetPct}{68.3\%}
\newcommand{\EwcRetPct}{59.1\%}

\newcommand{\FwdClampRetPeak}{0.797}
\newcommand{\FwdDoobLiftPts}{12.9}
\newcommand{\FwdDoobP}{0.004}
\newcommand{\FwdDoobSigStar}{0.6}
\newcommand{\FwdDoobWord}{an inverted-U}
\newcommand{\FwdMinSigStar}{0.05}
\newcommand{\FwdNoclampRet}{0.494}
\newcommand{\FwdOuLiftPts}{0.0}
\newcommand{\GateFLiftPts}{10.9}
\newcommand{\GateFPeakPct}{68.3\%}
\newcommand{\GateFPhi}{0.004}
\newcommand{\GateFPzero}{0.004}
\newcommand{\GateFSigStar}{0.02}
\newcommand{\GateFZeroPct}{57.4\%}
\newcommand{\KappaOneLiftPts}{13.0}
\newcommand{\KappaZeroLiftPts}{0.0}
\newcommand{\MesuRetPct}{65.2\%}
\newcommand{\NaiveRetPct}{57.4\%}
\newcommand{\NoiseTaxHiPct}{41\%}
\newcommand{\NoiseTaxLoPct}{4\%}
\newcommand{\NoiseTaxPct}{23\%}
\newcommand{\NumSeeds}{8}
\newcommand{\NumSigma}{10}
\newcommand{\OuRetPct}{57.4\%}
\newcommand{\OursMinusRFPts}{3.1}
\newcommand{\ReplayBuffer}{250}
\newcommand{\ReplayMinusOursPts}{14.6}
\newcommand{\ReplayRetPct}{83.0\%}
\newcommand{\SilTrainDoobAvgPct}{58.1\%}
\newcommand{\SilTrainDoobRet}{69.6\%}
\newcommand{\SilTrainDoobTaskOne}{46.6\%}
\newcommand{\SilTrainOuAvgPct}{59.0\%}
\newcommand{\SilTrainOuRet}{54.0\%}
\newcommand{\SilTrainOuTaskOne}{64.0\%}
\newcommand{\SilTrainRetGainPts}{15.6}
\newcommand{\SilTrainWallMin}{79}
\newcommand{\SiliconAdditivePct}{94.6\%}
\newcommand{\SiliconCVMaxPct}{12.0\%}
\newcommand{\SiliconCVMinPct}{1.6\%}
\newcommand{\SiliconChip}{hxcube7fpga3chip61\_1}
\newcommand{\SiliconMeasLiftPts}{13.2}
\newcommand{\SiliconMeasSigStar}{0.02}
\newcommand{\SiliconMeasWord}{survives}
\newcommand{\SiliconMultPct}{5.4\%}
\newcommand{\SiliconNumSendsExp}{0.47}
\newcommand{\SiliconNumSendsPts}{4}
\newcommand{\SiliconRepeats}{128}
\newcommand{\SiliconSignalRange}{8.2\times}
\newcommand{\SiliconTrialStd}{1.24}
\newcommand{\YYLiftPts}{7.4}
\newcommand{\YYSigStar}{0.035}
\newcommand{\YYSurviveWord}{reproduces}